# Enhancing Artificial intelligence Policies with Fusion and Forecasting: Insights from Indian Patents Using Network Analysis

Akhil kuniyil[1], Avinash kshitij[1,2*] and Kasturi Mandal[1]

[1] CSIR-National Institute of Science Communication and Policy Research, Dr. K.S. Krishnan Marg, New Delhi 110012, India
[2] Centre for Studies in Science Policy, School of Social Sciences, Jawaharlal Nehru University, New Delhi 110067, India

**Abstract**— This paper presents a study of the interconnectivity and interdependence of various Artificial intelligence (AI) technologies through the use of centrality measures, clustering coefficients, and degree of fusion measures. By analyzing the technologies through different time windows and quantifying their importance, we have revealed important insights into the crucial components shaping the AI landscape and the maturity level of the domain. The results of this study have significant implications for future development and advancements in artificial intelligence and provide a clear understanding of key technology areas of fusion. Furthermore, this paper contributes to AI public policy research by offering a data-driven perspective on the current state and future direction of the field. However, it is important to acknowledge the limitations of this research and call for further studies to build on these results. With these findings, we hope to inform and guide future research in the field of AI, contributing to its continued growth and success.

## I. INTRODUCTION

Artificial Intelligence (AI) technology has rapidly emerged as a critical tool in addressing the complex challenges faced by society in the modern era. AI has the potential to revolutionize a wide range of industries from healthcare and finance to transportation and agriculture [1] last but not least environmental hard and societal changes [2].

With the ability to analyze vast amounts of data and automate tasks that were once exclusively performed by humans, AI is reshaping the way we live and work. Given the potential of AI, it is essential to study and under-stand its applications, fusion of technologies, changes over the years in the domain as well as societal impacts. This un-derstanding is crucial for policymakers, as they must develop effective policies that keep pace with the rapid advancement of AI technology. Moreover, the study of AI is also rele- vant for individuals, businesses, and organizations, as they must be prepared to adapt to the changes brought about by AI. The study of AI is crucial in today's era to unlock the full potential of this groundbreaking technology and to address the challenges and opportunities it presents.

Artificial intelligence (AI) is increasingly driving impor- tant developments in technology, predicting an epidemic, and improving industrial productivity, since artificial intelligence emerged in the 1950s, the global patent filings in artificial intelligence (AI) recorded the highest average annual growth rate (AAGR) of 28%, reflecting it as the most prolific tech- nology [3]. Most of the AI patent filings were focused on machine learning (ML) models, speech recognition, image analysis, and natural language processing systems reveal- ing Global Data, a leading data, and analytics company [3]. Moreover, the ratio of scientific papers to inventions has de-creased from 8:1 to 3:1 in (2010-16) [4] indicative of a shift from theoretical research to the use of AI technologies in commercial products and services.

According to **Kazuyuki Motohashi**, *a Japanese corpo- rate economist at the University of Tokyo, while "science plays a significant role in the development of AI technology, the private sector's role becomes increasingly important as it gains ownership of intellectual property rights for inven- tions"*. This ratio of scientific papers to inventions serves as justification for this statement.

Some areas of AI are growing more quickly than others (eg: healthcare [5] ); Machine learning is the dominant AI technique disclosed in patents and is included in more than one-third of all identified inventions [4]. The field of Artifi-cial Intelligence technology is undergoing a revolution with the advent of machine learning techniques, particularly deep learning and neural networks. These techniques have seen staggering growth in terms of patent filings, with deep learn-ing leading the charge with an average annual growth rate of 20 to 23080 patent filings from 2012 to 2021. Among AI functional applications, those with the highest growth rate in patent filings were AI for robotics and control meth- ods, both of which saw an average growth rate of 55% per year. This highlights the significant impact and potential of these cutting-edge AI techniques in driving innovation and progress [4].

The growth rates observed in the identified AI-related patent data are notably higher than the average annual growth rate for patents across all domains of technology. From 2010 to 2021, the number of AI patents filed has grown exponen-tially, with the number of patents filed in 2021 being more than 30 times higher than in 2015, showing a compound an-nual growth rate of 76.9% [6]. The AI Index 2022 Annual Report, Stanford University, data highlights the rapid ad- vancement and increasing importance of AI technology [6].

Artificial intelligence (AI) patent studies are crucial for fu-ture policy making. They provide valuable insights into the state of the field and the direction of AI development. By analyzing patent data, policy makers can gain a better under-standing of which assignee, organizations, and countries are leading the way in AI research and development, as well as the domain, and sub-technologies, of AI technologies that are being developed and patented. This information can inform policy decisions and shape the future of the field.

According to a report by the International Association for AI and Law (IAAI), the number of AI-related policy pa- pers published by governments and international organiza- tions has increased significantly in recent years [7], and the growth in AI policy research and patent applications, there has also been a surge in the number of AI-related startups and investments in the field. Data from Crunchbase shows that the number of AI startups has more than doubled since 2015, and the amount of venture capital invested in AI has in- creased by over 300% in the same time period [8] across the globe. The combined growth in AI policy research, patent applications, startups, and investments in the field highlights the importance of AI and the need for responsible develop- ment and use.

The goal of this study is to enhance AI policy research by delving into the patenting activities, technology fusion, and maturity level of the technology, with a specific focus on the Indian context. *Technology fusion refers to the blur- ring of boundaries between disjoint areas of science, and technology. It is a phenomenon that is described by several terms, including technology convergence, merging, cross- fertilization, and hybridization* [9, 10] Our objective was to facilitate communication between the fields of AI technolo-gies and policy making by delving into the technical aspects of patent data. We will conduct a detailed examination of the following components to achieve this goal.

- Examine the prevalent trends in AI patents and their evolution over time.

- Analyze how different technologies (International patent classification code *IPC*) converge and fuse in the n field of AI, and the impact of this convergence on the development of AI.

- Analyze how different technologies (International patent classification code *IPC*) converge and fuse in the field of AI, and the impact of this convergence on the development of AI.

- Investigate how the growth, maturation, and saturation of AI technologies affect the formation of a digital ecosystem.

- Evaluate the implications for policymakers and stake- holders of the formation of a digital ecosystem in re- lation to the growth, maturation, and saturation of AI technologies.

In order to accomplish our objectives, we employed social network analysis in tandem with patent data. This enabled us to investigate the connections and interactions between vari-ous actors in the technology domain. Network analysis is a way of thinking about the complex system that focuses on the relations among the system entities. Patent network analysis is one of the popular areas of social network analysis (SNA). *Social network analysis refers to a set of methods used to analyze relational data from a structural perspective, as de-fined by John Scott* [11]. Various network-based metrics and graph representations enable researchers to comprehend the patent network structure and its interactions. patents are one of the best indicators of science and technology transfer [12]. Patenting technology is typically the preliminary step that ev- ery technology developer takes, and this allows us to connect science to technology and technology to society in the form of products or services.

## II. INDIAN CONTEXT

India has made significant contributions to the field of arti- ficial intelligence (AI) and has a growing presence in AI re-search studying the patent network in the field AI in India can potentially be beneficial for both developed and developing countries There are several statistics and pieces of evidence that demonstrate India's growing presence in the field of arti-ficial intelligence and its contributions to AI research. Some potential reasons why you might choose to focus on Indian patents or patents of Indian origin in your case study could include the following:

According to a report by the National Association of Soft-ware and Services Companies (NASSCOM) [13], the AI market in India is expected to grow to $16 billion by 2025,
[13] driven by the increasing adoption of AI in various sec- tors such as healthcare, banking, and retail.

The Indian government has also recognized the impor- tance of AI and has taken steps to promote the development and adoption of AI in the country [14]. For example, the gov-ernment has launched the "National Artificial Intelligence Portal," which aims to provide a platform for researchers, developers, and users to come together and share knowledge and resources related to AI [14].

There are many AI research organizations and academic institutions in India that are actively engaged in research and development in the field. For example, the Indian Institute of Technology (IIT) has several research groups and centres focused on AI, including the Center for Artificial Intelligence and Robotics and the Machine Learning and Data Science Group [15]. India has also produced many successful AI startups in recent years, such as Haptik [16, 17], which provides AI-powered chat-bot services, *Niki.ai-* an AI-powered chat-bot that helps users make online purchases, book cabs, and pay bills [18] and *Flutura* an IoT analytics company that uses AI to optimize industrial processes [17]. These companies are evidence of the country's growing entrepreneurial spirit in the field of AI [19].

## III. RELATED WORK

### a. Innovations in Fusion Technology: A Literature Review

In the related work of technology fusion, two main research streams can be distinguished: those that focus on theoretical aspects and those that conduct empirical case studies for field-level data analysis. Studies in the first stream, such as those by Kodama [20] and Hacklin [21], focus on the theoretical definition of convergence. Curran [10] specifically organizes work on convergence by defining it as "a blurring of boundaries between at least two hitherto disjoint areas of science, technology, markets or industries, it creates a new sub-segment as a merger of /parts of the old segments" (p. 22) [10]

This study uses the terms "convergence" and "fusion" interchangeably but highlights their difference if relevant to the analysis. Curran [10] defines fusion as the "merger of /parts of the old segments" in the very same place of at least one of the objects. Similarly, Pennings and Puranam [22] in their analytical framework distinguish between demand-side and supply-side convergence. The latter is related to technological functionality, the former is associated with the contemporary satisfaction of different needs based on different technological capabilities which converged to become similar.

Several authors have built [23, 24, 25] upon the seminal study of Pennings and Puranam [22] to further explore the concept of convergence in technology fusion. Bröring [26] identified additional categories such as "technology-driven input-side convergence" and "market-driven output-side convergence" which evolved from new technologies applied across different industries.

This research focuses on the "technology" locus, which corresponds to the supply side in Pennings and Puranam's [22] framework. The motivation for this research comes from the work on technology convergence highlighted by Kim and Kim [27] and Caviggioli, and Federico [9]. Additionally, the scarcity of data on the convergence between two technologies and the absence of a widely accepted indicator of inter-disciplinarity limit the analysis of technology fusion [28, 27]. The present study endeavours to unveil a novel method for identifying technology convergence at the system level utilizing patent data and further delves into analyzing the key drivers of technology fusion, with a specific emphasis on Artificial Intelligence. (The main contribution of this study is to introduce a unique method for identifying technology convergence at the system level using patent data and to investigate the factors driving the fusion of technology, specifically Artificial Intelligence.) To summarize, the technology fusion of AI using patent data is a rapidly evolving field with many exciting research directions. By leveraging the wealth of information contained in patent data, researchers are working to develop and improve AI technologies in a way that is more efficient, effective, and impactful.

## IV. METHODOLOGICAL FRAMEWORK

Our methodology consisted of several steps to analyze the evolution and fusion of AI technology Figure. Firstly, we used the Gompertz equation to model the growth of AI technology using patent data. This allowed us to forecast the future growth of the technology and gain insights into its maturity level. The detailed methodology for this analysis is included in the conceptual framework. Next, we created a co-occurrence matrix of IPCs to identify the most important technologies within the dataset. By analyzing the frequency of co-occurrence between different IPCs, we were able to understand which technologies were driving the most innovation during each time period. We then quantified the importance of each technology using centrality measures, such as degree centrality and betweenness centrality. We also examined the degree of fusion of IPC sections, classes, and sub-classes to understand how different technologies were being integrated into new products and services. By analyzing the degree of fusion, we were able to better understand the relationships between different technologies and how they were contributing to overall technological development. lastly, we conducted a four-time frame analysis to track the growth of the International Patent Classification (IPC) system over time. By analyzing the growth of IPC over the four-time frames, we were able to understand how different technologies were evolving and how their importance changed over time. This helped us to identify any trends or patterns that emerged and to understand the overall direction of technological development. The detailed methodology for this analysis is included in the conceptual framework.

### a. Conceptual frameworks

#### 1. Patent analysis

It has been a critical issue to understand technological trends not only to avoid unnecessary investment but also to gain the seeds for technological development. So, many methods have been developed to recognize the progress of technologies, and one of them is to analyse patent information. However, it is hard for non-specialists to analyse patent information because patent information is enormous and rich in technical and legal terminology. Therefore, patent information needs to be transformed into something simpler and easier to understand and visualization methods are considered to be proper for representing patent information and its analysis results. Generally, visualization methods are known as one of the best data mining ways to understand because graphical display methods often offer superior results com- pared to other conventional techniques [29, 30].

This study applies a social network analysis (SNA) methodology to obtain and interpret the technology fusion characteristics of AI technologies included in their respectivenetworks. Network analysis has been widely used for inter-preting complex systems from multiple perspectives, rangingfrom social systems to shared knowledge networks [29]. Thevisual image support bringing out the internal connections between individual nodes that form the network. In SNA, links show relationships or flow between the nodes. SNA focuses on the relations among actors, not individual ac- tors, and their attributes. This methodology has been widelyused to understand the complicated interactions in techno- logical evolution since the network structure of patents and their IPCs can explain the complicated inter-dependency andtrends in technological development and fusion [24].

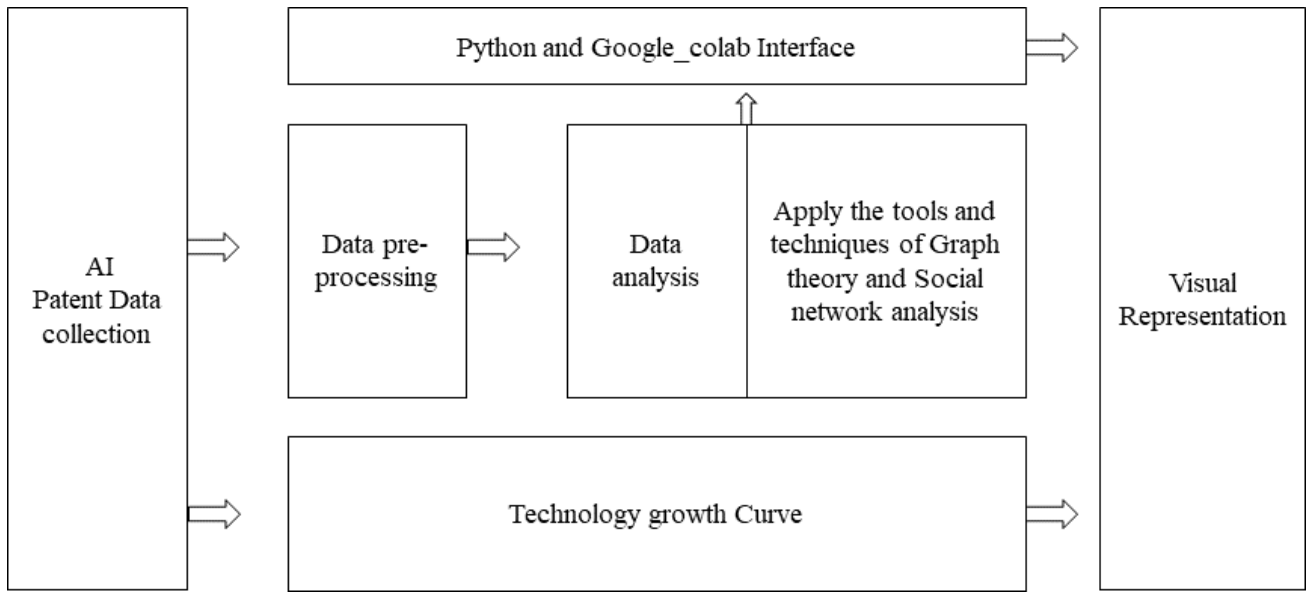

**Fig. 1:** Schematic Representation of Methodology through Block Diagram

## 2. Patent Classification System

A patent classification system is an arrangement of hierarchi-cal categories for technology codes. The primary purpose isto assign a patent examiner administratively and support effi-cient patent searches by arranging patent documents system-atically [31]. There are several patent classification systems.The International Patent Classification (IPC) and the Coop- erative Patent Classification (CPC) are common and used bypatent offices worldwide including the USPTO, WIPO, and EPO. We focus on IPC-based analysis.

The International Patent Classification (IPC) is a hierarchi- cal taxonomy developed and administered by the WIPO for classifying patent documents. The IPC covers a wide range of technical and scientific vocabulary. The IPC comprises eight sections from A to H, which are in turn subdivided intoclasses, sub-classes, groups, and sub-groups, and regularly revised to include new technologies, or the existing classifi-cation is divided into several sub-units with a more narrowlydefined scope. Data analysts seek to gain a deeper under- standing of the special structures, knowledge, and economicvalues that underlie patent data and close the gap between big data analytics capacities. International Patent Classification (IPC) system [31] under which each patent can be broadly classified under one (or more) of the eight classes or sectorsfrom A to H.A: Human Necessities

- B: Performing Operations; Transporting

- C: Chemistry; Metallurgy

- D: Textiles; Paper

- E: Fixed Constructions

- F: Mechanical Engineering; Lighting; Heating;Weapons; Blasting

- G: Physics

- H: Electricity

## 3. Network Structural Indicator

Centrality measures are mathematical algorithms that are used to identify the most important nodes in a network. These measures can be used to quantify the fusion of technol-ogy by identifying the technologies that are driving the fusion process and are likely to be adopted more widely in the fu- ture. There are several different centrality measures that canbe used in this context. Perhaps the most popular and widelyused measure to identify central positions and power in a net-work [32, 33, 34]. It can be examined with a variety of van-tage points, such as "does a node shows the highest number of connections to others," or "is a node frequently interposedbetween other nodes," or "is there a node that has the shortestpath length to all other nodes?" Each question offers a uniqueperspective on centrality degree centrality, betweenness cen-trality, and closeness centrality.

Degree centrality $C_D(i)$ is a baseline metric of connected-ness, which refers to the number of direct edges a node has. The assumption with degree centrality is that the number of connections is a key measure of importance within a network[33, 34].

## IV METHODOLOGICAL FRAMEWORK

### a Conceptual frameworks

IPC code co-occurrence structure reveals connectivity and relations among the distinguished technology areas. More-

Degree centrality $C_D(i) = \sum_{}^{n} G_{ij} = \sum_{}^{n} G_{ji}$ for $i = 1, 2, \cdots, n$

over, an IPC code that has a high centrality value in a net-work can be considered to be a core and main technologyarea [24]. The derivation process for IPC co-occurrence net-

$G_{ij}$ represents the value of the link between node i and node j (the value is either 0 or 1); and n: represents the num-ber of nodes in the network Power and placement are funda-mental to network analysis. Degree centrality quantifies the total connectivity of a node, that is, power. However, it doesnot specify a node's location in a network (i.e., placement) In other words, when computing degrees in an ego network,degree centrality focuses on the focal nodes and ignores all the other nodes in the network. Contrastingly, betweenness $C_B$ (i) and closeness centrality Cc consider the complete net-work. Betweenness centrality $C_B$ measures the number of shortest paths that pass through a node [33, 34]

$Betweenness\ Centrality\ C_B(i) = \sum \frac{G_{jk}(i)}{}$ for $i \neq j \neq k$

works, where nodes are defined by IPCs and links are defined by the co-occurrence of IPCs in a patent. This assumes that if a certain IPC code co-occurs with another, there is a closerelationship between the technology areas and they can be considered to be linked. Forming a network with individual IPC codes as nodes has the advantage of analyzing the tech-nology level over the patent-cited level network.

<κ(2)

$G_{jk}$: represents the number of shortest paths linking nodesj and k; and $G_{jk}(i)$: represents the number of the shortest paths linking nodes j and k that pass through node i. Be- tweenness centrality is regarded as indicating how much po-tential control a node has over the flow of information. Be- tweenness centrality relates to questions such as "who is crit-ical for information flow" or "who plays the role of interme-diary or broker in a network?" While betweenness centralityfocuses on the role of a brokerage or gatekeeper of informa-tion flow in a network, closeness centrality $C_c$ is considered a global metric closeness centrality $C_C$ cores each node based on its closeness to all other nodes in the network. A node with high closeness centrality can quickly reach out to othernodes without relying much on intermediaries in the network[33, 34, 35].

Closeness Centrality $C_c(i) = \frac{1}{\sum}$ for $i \neq j$ (3)

| | IPC 1 | IPC 2 | IPC 3 | IPC 4 | ..... | IPC n |
|---|---|---|---|---|---|---|
| IPC 1 | 0 | 1 | 0 | 2 | | n1 |
| IPC 2 | 1 | 0 | 2 | 1 | | n2 |
| IPC 3 | 0 | 2 | 0 | 3 | | n3 |
| IPC 4 | 2 | 1 | 3 | 0 | | n4 |
| ... | | . | . | . | . | n... |
| IPC n | .. | .. | .. | .. | .. | nm. |

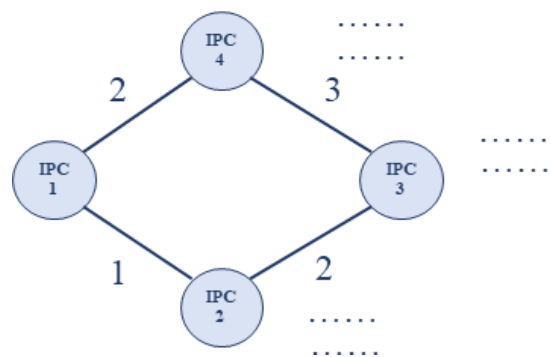

## 4. Co-Occurrence Network

The centrality measures can provide valuable insights into the structure and organization of an IPC code graph and help
The co-occurrence matrix, a powerful tool in patent analysis for identifying important technology areas within a domain, is derived from the co-occurrence network Figure 2. It maps the frequency of occurrence of keywords or concepts in a set of patents and helps in uncovering the relationships between them. By analyzing the co-occurrence patterns, researchers can gain insights into the technological trends, research focus, and potential opportunities for innovation in a particular field. In this context, the IPC (International Patent Classification) system serves as a standardized taxonomy to classify patents and facilitates the identification of related technology areas. Depending on the depth of analysis, different levels of IPC codes can be applied. In this study, network analysis has been conducted at the main-group level, since this level has been widely accepted by previous studies [24].

In general, a patent contains at least one IPC code. However, if a patent contains more than one IPC code, it can be assumed that multiple technology areas have been converged and integrated within the patent [24]. Thus, comparing the to identify the most central or influential codes within the graph. These measures can be useful for understanding the technical content of a research or patent application and iden-tifying key areas of focus or expertise.

## 5. Degree of Fusion $D_F$

In simpler terms, technology convergence or fusion refers to when different technologies overlap or come together to create something new. This is often valuable because it allows for interdisciplinary research that can lead to innovative solutions [9]. There are several ways to measure the fusion of technology using patent IPC (International Patent Classification) codes and supporting data. One approach is to calculate the degree of fusion, which is a measure of the extent to which technologies are combined or integrated in a patent.

To use this formula, we need to obtain the IPC codes for the patent (or patents) that you are analyzing. We can then count the number of unique IPC codes and the total number of IPC codes in the patent (or patents), and use these values to calculate the degree of fusion [36],

$$Degree\ of\ Fusion\ D_F = \frac{\sum_{i=1}^{n}}{\sum_{j=1}^{n}}$$

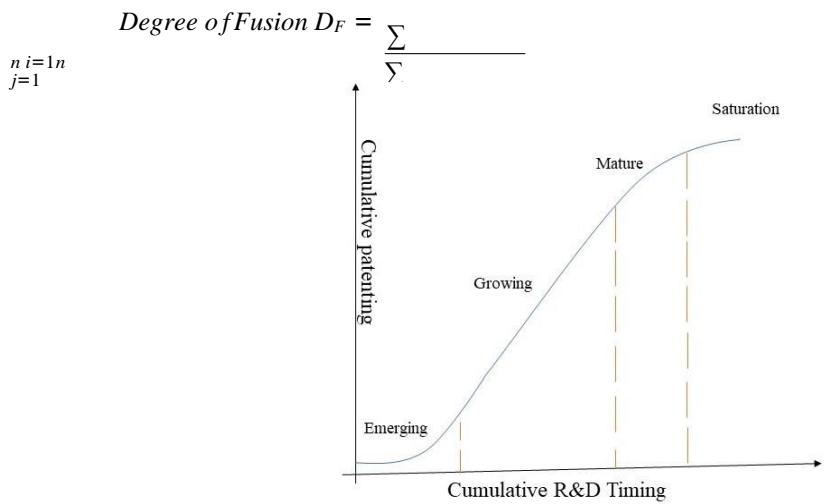

$$IPC_u(i) IPC_t(j) \quad (4)$$

Where $IPC_u(i)$ represents the number of unique IPC in that section/ subsection in the set of patents, $IPC_t(j)$ represents the total number of IPC code in the set of patents.

## 6. Technology life cycle TLC

One way to use patent data to study the technology life cy- cle is to analyze the trends in the number of patents filed over time for a particular technology or group of technolo- gies [37]. This can give an idea of the level of innovation and development activity in the field and can help to iden- tify key periods of growth or decline Using patent data to study the technology life cycle can provide valuable insights into the evolution and current state of a particular technology, and can help to make informed decisions about investment, research, and development [38, 39].

The Gompertz curve is a mathematical model that is often used to describe the adoption or diffusion of new technolo- gies over time. The curve has an "S" shape, with a slow initial adoption phase followed by a rapid growth phase and then a slow decline phase Figure 3. The curve is used to fit the S-curve of patent growth, which is found appropriate for forecasting technology growth. In general, the Gompertz and Logistic models are the most widely adopted approach for fitting the S-curve [40] and have better performance than the existing models [41]. To select the optimal fitting model, we adopt the R square value, root mean squared error (RMSE), and mean absolute percentage error (MAPE) to test the fitting effects [42, 43, 44]. Testing results indicate that the Gom- pertz curve fits better than the Logistic model. Therefore, we use the Gompertz curve to calculate the technology ma- turity. The Gompertz curve is a mathematical model that is often used to describe the adoption or diffusion of new tech-nologies over time. The curve has an "S" shape, with a slow initial adoption phase followed by a rapid growth phase and then a slow decline phase. The curve is used to fit the S-curve of patent growth, which is found appropriate for forecasting technology growth. In general, the Gompertz and Logistic models are the most widely adopted approach for fitting the S-curve [40] and have better performance than the existing models [41]. To select the optimal fitting model, we adopt the R square value, root mean squared error (RMSE), and mean absolute percentage error (MAPE) to test the fitting ef-fects [42, 43, 44]. Testing results indicate that the Gompertz curve fits better than the Logistic model. Therefore, we use the Gompertz curve to calculate the technology maturity.

The formula presents how to calculate the Gompertz curve [40].

$$Y_t = Le^{-ae^{-bt}} \quad (5)$$

It depends on the three coefficients L, a and b, Where a and b determine the location and shape of the curve respectively and L denotes the asymptotic value. Further, this value is calculated using the ordinary least square value method to solve three unknown variables.

**Fig. 3:** Assessing Technological Maturity using the Gompertz Curve, a type of S-curve

## V. RESULTS AND DISCUSSION

The present study aims to highlight the significance of utiliz-ing patent analytics as a tool for informed decision-making in the public policy-making context. By incorporating inter-disciplinary viewpoints, the findings of this research demon-strate the potential of network analysis to provide a deeper understanding of complex systems through the examination of relationships between entities within the system [34]. Our research on the evolution and fusion of artificial intelli-gence technology revealed a number of interesting findings. We found that the overall number of AI-related patents in- creased significantly over the ten-year period, with the great-est growth occurring in the most recent time frame. This sug-gests that AI technology is currently experiencing a period of rapid development and innovation.

We also found that certain subcategories within the IPC classification system, such as machine learning and natural language processing, showed particularly strong growth over the ten-year period. This highlights the increasing impor- tance of these specific areas within the field of AI.

As the first result, The maturity level of a technology can be an important indicator of its potential impact and adop- tion. In this study, we assessed the maturity level of the tech-nologies under investigation to gain insight into their current state of development and future trajectory. In this study, we used the Gompertz curve to model the growth patterns of the technologies under investigation and determine their matu- rity levels.

### a. Technology Maturity level

The Gompertz curve is a commonly used mathematical model in technology analysis that describes the growth of technology over time. We fitted the Gompertz curve to the patent data of each technology and used the curve parameters to estimate the maturity level. The curve is a mathematical model that is often used to describe the growth of a system over time (based on a Non-linear equation)

Figure 4 includes two graphs. The first graph depicts a comparison between the predicted and actual growth of AI technology from 2012 to 2022. The results indicate that the predicted and actual data are closely aligned throughout the entire period, with the predicted data slightly surpassing the actual data from 2020 to 2022. These findings suggest that the Gompertz model is a good fit for the data, although the actual growth of AI technology may be slightly higher than what was initially predicted.

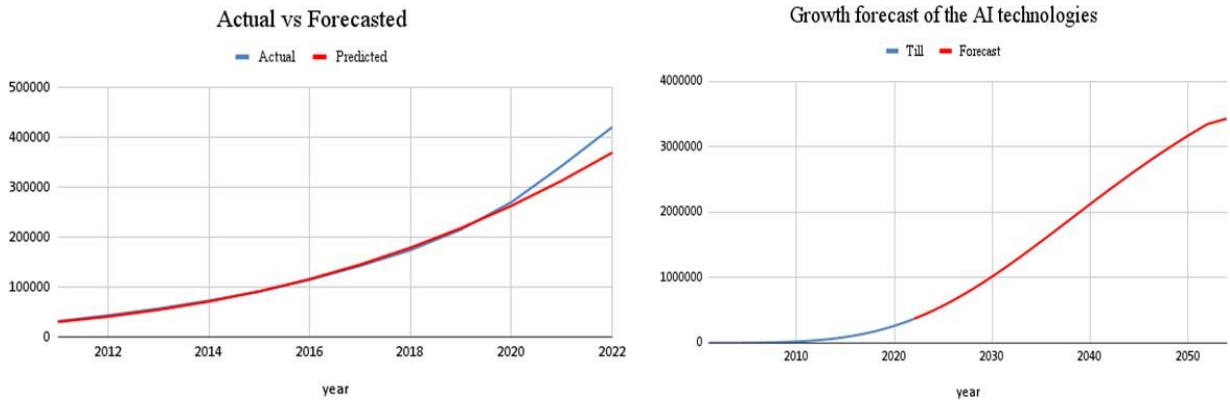

**Fig. 4:** Predicted Growth in Technological Maturity for AI Technology

The second graph displays a single curve that represents the predicted growth of AI technology. The results suggest

| IPC | Degree | IPC | Betweenness | IPC | Closeness | IPC | Clustering Coefficient |
|---|---|---|---|---|---|---|---|
| G06F | 165 | G06F | 0.127195 | G06F | 0.386941 | G07F | 1 |
| G06K | 87 | G06T | 0.091007 | G06K | 0.367923 | G07D | 1 |
| G06T | 95 | G06K | 0.063525 | G06N | 0.361097 | B29C | 1 |
| G06N | 91 | G05B | 0.056241 | H04L | 0.353233 | G06G | 1 |
| H04L | 90 | G06F | 0.055328 | G06Q | 0.352593 | A01G | 1 |
| G06Q | 77 | H04L | 0.034848 | G06F | 0.340265 | F01D | 1 |
| G05B | 71 | G06Q | 0.02593 | G10L | 0.33157 | G10K | 1 |
| H04N | 71 | G05B | 0.020349 | G06K | 0.324927 | A63F | 1 |
| G10L | 58 | G05B | 0.020259 | G06T | 0.322772 | F01K | 1 |

that the technology is still in the growth phase and has not yet reached a mature level. The curve demonstrates a gradual in-crease in the growth rate until 2020, followed by a significant acceleration in the growth rate, indicating that the technology is currently experiencing a period of rapid expansion. It is important to note that the curve predicts a gradual saturation point after 2050, which suggests that the technology will take some time to reach its full potential. These findings provide support for the suitability of the Gompertz model in captur- ing the growth pattern of AI technology and indicate that the technology has a promising future ahead of it as it continues to develop and expand.

It is worth noting that patents are typically classified based on their technologies rather than their products or industries[5]. As such, an important next step would be to capture the significant technologies within the AI domain. This can be achieved by quantifying the importance of IPC technologies using a co-occurrence matrix and centrality measures. This approach would enable us to identify the key technologies and their interrelationships within the AI domain and pro- vide further insight into the growth and development of this rapidly expanding field.

## b. Core technologies correspond to its centrality

The co-occurrence matrix of technology classification was analyzed in our study, yielding several key insights into the interconnections between different artificial intelligence technologies. The results of this analysis are presented in Table 1.

To evaluate the significance of each technology in the field of AI, we first created a co-occurrence matrix of all tech- nologies (IPC) and selected the top 10 based on their central-ity measures, including degree centrality, betweenness cen- trality, closeness centrality, and clustering coefficient, which were calculated based on their frequency of occurrence and connections to other technologies in the co-occurrence ma- trix presented in Table 1. Degree centrality indicates the ex-tent to which technology dominates the field, with a higher degree denoting a greater presence in the network. Con- versely, betweenness centrality serves as a valuable tool for assessing the role of technology in connecting different sub-domains of AI. A higher betweenness centrality score indi- cates a technology's ability to act as a bridge between subdo-mains and facilitate knowledge transfer.

| | | | | | | G05D | 31 |
|---|---|---|---|---|---|---|---|
| G05D | 0.013281 | H04N | 0.318546 | G01C | 0.833333 | | |

**TABLE 1:** MEASURING THE SIGNIFICANCE OF TECHNOLOGY THROUGH CENTRALITY METRICS

In order to assess the significance of each technology in the AI field, we calculated centrality measures, including de-gree centrality, betweenness centrality, closeness centrality, and clustering coefficient. The top 10 technologies are pre- sented in Table 1. The degree centrality measure reflects the dominance of technology across the domain and a higher de-gree indicates a greater presence in the network.

On the other hand, the betweenness centrality measure is a valuable tool for evaluating the role that technology plays in bridging different subdomains of AI. The betweenness cen- trality score reflects the extent to which technology acts as a connector between different subdomains and facilitates the flow of knowledge spillovers from one subdomain to another. In our analysis, the technologies classified under IPC sub- classes G06F (Electric Digital Data Processing), G06T (Im-age data processing or generation), G05B (Remote control systems), and G06K (Recognition of data; Presentation of data; Record carriers; Handling record carriers) have been identified as having high betweenness centrality scores, in- dicating their crucial role in bridging subdomains of AI and facilitating knowledge transfer.

The closeness centrality measure captures the commonal-ity of technology in all AI patents, with technologies hav- ing a high closeness centrality score considered as crucial and fundamental components shaping artificial intelligence across all sectors.

The evaluation of closeness centrality in our research aimed to assess the widespread' prevalence of technology within the entire AI patent landscape. The outcome of this analysis determined the cruciality and fundamental nature of the technologies to the development and advancement of ar-tificial intelligence across various industries. The analysis found that the technologies classified under the International Patent Classification (IPC) subclasses G06F, G06K, G06N, and H04L, in particular, demonstrated high levels of close- ness centrality.

The IPC subclass G06F encompasses the area of electric digital data processing and includes topics such as comput- ers, data processing methods, and software. G06K pertains to the recognition of data, including character recognition and image analysis, which is crucial for AI applications in image and speech recognition. G06N covers data process- ing systems and methods for control, measuring, and testing, making it an essential component for the development of AI across various industries. And, H04L addresses the transmis-sion of digital information, including the transmission of dig-ital signals, digital communication, and digital information processing, making it vital for the development of AI systems and applications. Finally, from Table 1, we calculated the clustering coefficient to gain insight into the extent to which technologies tend to form clusters or communities. The clus-tering coefficient measures the degree to which nodes in a network tend to cluster together or form cliques. A higher clustering coefficient indicates a greater tendency for nodes to cluster together, suggesting the presence of distinct subdo-mains within the AI field. Our findings indicate that the. The clustering coefficient is a valuable tool for identifying areas of specialization and expertise within the field and gives a better understanding of the structure of the AI research land- scape. For a more in-depth analysis of the interconnectivity and interdependence between various AI technologies, we introduce the degree of fusion measure. This measure re- flects the number of connections and relationships between the technologies in a comprehensive manner and provides a clear understanding. The degree of fusion result is a key fac-tor in identifying the crucial areas of fusion in the field of AI.

## c. Analysis of the fusion of degree

Our analysis of the fusion of degree within different IPC classes revealed a number of interesting findings. We found that certain IPC classes, such as G06 and G10, had a higher fusion of degree values than others, indicating that technologies within these classes were frequently being integrated into new innovations Table 2. On the other hand, other IPC classes, such as B33 and C09, had a lower fusion of degree values, indicating that technologies within these classes were less commonly being fused into new innovations.

Based on the results provided, it appears that the degree of fusion was calculated for each IPC class within the categories listed. The IPC codes G06, G10, H04, H01, B06, B33, A61, A63, C12, C09, F24, and F01 correspond to specific subcategories within the technology domains previously mentioned.

The IPC code G06 refers to computer technology and includes subcategories such as computer hardware, software,

| IPC section corresponding to degree of fusion | IPC class corresponding to degree of fusion | IPC sub class correspond to degree of fusion |
|---|---|---|
| G : 0.69 | G06 : 0.044<br>G10 : 0.042<br>G01 : 0.029<br>G05 : 0.029<br>G08 : 0.014 | G06F : 0.245<br>G06N : 0.145<br>G06K : 0.117<br>G06T : 0.078<br>G06Q :0.066 |
| H : 022 | H04 : 0.180<br>H01 : 0.008<br>H02 : 0.003<br>H03 : 0.002<br>H05 : 0.001 | H04L : 0.073<br>H04N : 0.050<br>H04W : 0.020<br>H04M : 0.010<br>H04L :0.005 |
| B : 0.036 | B60 : 0.011<br>B33 : 0.005<br>B64 : 0.004<br>B29 : 0.004<br>B07 : 0.002 | B60R : 0.004<br>B33Y : 0.004<br>B60W : 0.003<br>B64D : 0.002<br>B07C : 0.001 |
| A : 0.036 | A61 : 0.226<br>A63 : 0.002<br>A01 : 0.001<br>A23 : 0.0007<br>A47 : 0.0007 | A61B : 0.001<br>A63F : 0.001<br>A61K : 0.002<br>A61N : 0.002<br>A63H : 0.001 |
| C : 0.005 | C12 : 0.002<br>C09 : 0.001<br>C07 : 0.0005<br>C10 : 0.0005<br>C40 : 0.0005 | C12Q : 0.001<br>C12P : 0.001<br>C09D : 0.001<br>C09K : 0.001<br>C12N : 0.001 |
| F : 0.008 | F24 : 0.003<br>F01 : 0.001<br>F02 : 0.0007<br>F04 : 0.0005<br>F25 : 0.0005 | F24F : 0.003<br>F01D : 0.002<br>F01K : 0.002<br>F02C : 0.002<br>F25D : 0.001 |
| D : 0 | Not found | Not Found |

**TABLE 2:** IPC SECTION, CLASS, AND SUBCLASS INTEGRATION LEVEL (DEGREE OF FUSION)

and communication systems. The IPC code G10 refers to optical elements, systems, and instruments and includes subcategories such as lenses, prisms, and telescopes. The IPC code H04 refers to telecommunications and includes subcategories such as telephonic communication, radio communication, and television. The IPC code H01 refers to electricity and includes subcategories such as electric techniques and electric communication systems. The IPC code B06 refers to metallurgy and includes subcategories such as metalworking, metal casting, and metal production. The IPC code B33 refers to crystallography and includes subcategories such as crystal growth and crystal structure. The IPC code A61 refers to medical or veterinary science and includes subcategories such as medical instruments and devices, diagnostic methods, and surgery. The IPC code A63 refers to human necessities and includes subcategories such as hygiene and personal care products. The IPC code C12 refers to biochemistry and includes subcategories such as enzymes, hormones, and nucleic acids. The IPC code C09 refers to organic chemistry and includes subcategories such as chemical reactions and compounds. The IPC code F24 refers to heating and air conditioning and includes subcategories such as air conditioning systems and heating equipment. Finally, the IPC code F01 refers to machines and machine tools and includes subcategories such as engines, pumps, and motors.

The IPC Subclass, codes G06F, G06N, G06K, and G06T correspond to specific subcategories within the computer technology domain (G06). The code G06F refers to com- puter hardware and includes subcategories such as digi-tal computers, processors, and memory devices. The code G06N refers to data processing systems and includes subcat-egories such as database management, data mining, and datavisualization. The code G06K refers to information technol-ogy and includes subcategories such as image processing,

VI DISCUSSION

VII d Four Time windows where the Technologies evolved (as per degree)

VIII pattern recognition, and natural language processing. The code G06T refers to electrical communication systems and includes subcategories such as wireless communication andnetworking.

The IPC codes H04L, H04N, H04W, and H04M corre- spond to specific subcategories within the telecommunica- tions domain (H04). The code H04L refers to transmis- sion systems and includes subcategories such as antennas, modems, and multiplexing. The code H04N refers to digi- tal communication and includes subcategories such as error correction, data transmission, and data security. The code H04W refers to wireless communication and includes sub- categories such as satellite communication, cellular commu- nication, and wireless networking. The code H04M refers to telephonic communication and includes subcategories such as telephone systems, voicemails, and call centres.

The fact that these specific IPC classes have high degreesof fusion suggests that technologies within these subcate- gories are frequently being integrated into new innovationsand are playing a central role in driving technological devel- opment. For example, the high degree of fusion of G06Ftechnologies may indicate that advances in computer hard-ware are driving the development of new products and ser-vices. Similarly, the high degree of fusion of H04N technolo-gies may suggest that innovations in digital communicationare playing a key role in driving technological development.These results provide valuable insights into the relation- ships between different technologies and how they are in-fluencing one another and being integrated into new inno-vations. They also highlight the importance of certain IPCclasses in driving technological development and innovation.Next, we move on to our last result, where we present a comprehensive analysis of the core technologies, their de-gree of fusion, and their evolution over time in a quantitativemanner. By utilizing analytical tools, we were able to exam-ine the interplay between these technologies and identify thekey drivers of technological change within each time frame.Our findings shed light on the dynamic nature of technologi-cal progress and provide valuable insights into the future di- rection of technologies.

### d. Four Time windows where the Technologiesevolved (as per degree)

we have divided all IPC technologies into four distinct time frames to conduct a detailed analysis of their evolution overthe years Figure 5 This approach enabled us to gain a deeperunderstanding of how these technologies have progressed over time and identify key trends and patterns that have emerged within each time frame. The division of patent dataInto four-time frames allows for a more comprehensive anal-ysis of trends and developments within a particular field. Bylimiting the data to a single frame of 10 years, there is a riskof bias due to the release of new technologies or a sudden increase in patent applications, particularly those that are ac-companied by new IPC codes.

By dividing the data into four time frames, it becomes pos-sible to identify longer-term trends and patterns, rather than being influenced by short-term fluctuations. In addition, di- viding the data into time frames allows for a more accurate comparison between different time periods, as the data is not influenced by changes in the rate of patent applications or theintroduction of new technologies, dividing patent data intotime frames helps to provide a more accurate and unbiasedrepresentation of the evolution of a specific field [45, 46, 47].Figure 5 provides an illustration of the technological ad- vancements from 2012 to 2021 through four-time frames.The analysis of the International Patent Classification (IPC)group reveals the growth of artificial intelligence (AI)patents, with a significant increase from 52 patents in 2012to 335 patents in 2015. The most prevalent core technologyin 2012 was G06F17 (9 patents), which pertains to "elec-tronic digital data processing." This was followed by G06K9(technologies for manual recognition of written characteris-tics, patterns, fingerprints, etc.), G06F9 (recognition of data;presentation of data; record carriers), and H04L29 (transmis- sion of digital information, communication control, etc.).

From 2015 to 2018, there was a noticeable upsurge in thedemand for new technologies in image data processing and image analytics (G06T). The progression of this technologyfrom the last position in the first time frame to the third po- sition in the third time frame (67 degree) showcases a grow-ing demand for image data processing and visual data an- alytics. Certain technologies have consistently maintained their top positions over the years, including G06F11 (digitalcomputing/data processing, text processing, voice data anal-ysis/synthesis), H04L29 (transmission of digital information,communication control, etc.), and G06K9 (technologies for manual recognition of written characteristics, patterns, fin- gerprints, etc.) Among these, G06K9 has demonstrated con-sistent demand and relevance in the field, with the degreeof 7, 69, 194, and 575 in the four-time frames, respectively. These findings provide meaningful insights into the stability and growth of specific technologies in the field of artificial intelligence.

## VI. DISCUSSION

The discussion of our research paper is informed by the over-arching journey of our investigation. Our examination of the maturity level of AI technology has revealed that it has pro-gressed from the emerging stage and is now in a period of growth that is expected to continue for a considerable time before reaching saturation Figure 4.

We chose to use patents as an indicator for analyzing AI technology because patents are classified based on the un- derlying technology and not on the industry or products [5]. Analysis, we first constructed a co-occurrence matrix of IPC to identify frequently occurring technologies. We then ap- plied centrality measures to quantify the importance of each technology in Table 1. Next, we examined the degree of fu- sion among the technologies in Table 2. Lastly, we inves- tigated the core technologies in four different time frames to track their evolution over the years and identify the fo- cus area domination. The division into different time frames was made to avoid any bias and gain a comprehensive under- standing of the technological evolution over the years.

The examination of IPC over four distinct time frames has revealed that, during the period from 2012 to 2018, the bulk of patents were focused on data mining, information re- trieval, and data generation technologies. Nevertheless, since the start of 2020, there has been a shift in this trend towards applied AI technologies such as image recognition, genetic studies, and advanced pattern identification techniques. Ad- ditionally, the use of algorithms, especially support vector machines (SVM), has become widespread post-2020. This phase of growth is distinguished by a diverse range of appli- cations of machine learning problems in classification, facil- itating the efficient identification of AI's capabilities and its adaptability across various domains.

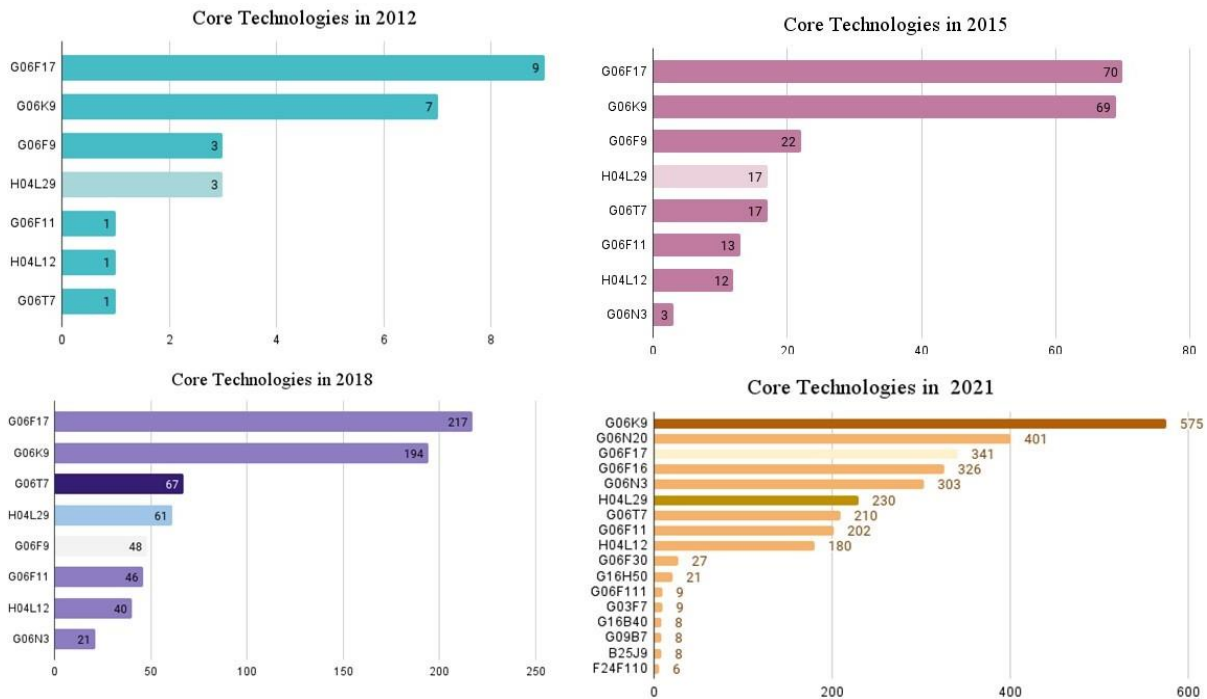

**Fig. 5:** IPC Evaluation over years, in the four-time frames

The growth of AI technology is marked by significant shifts in the focus and applications of this field. Today, when people talk about AI, they are mostly referring to machine learning, which has its roots in computer science dating back to the 1950s. However, the methods for building recommen- dation engines, spam classifiers, or traffic predictions that are popular today are fundamentally different from the algo- rithms of the past. The key difference lies in the availability of data. The proliferation of data sources such as GPS, im- ages, credit card purchases, and smart bands has made it clear that AI technology has the potential to significantly impact society. However, the challenge now is to effectively employ this collection of technologies in a human-centred way that benefits communities, businesses, and governments. Thus, the current conversation about AI is a complex one that re- quires careful consideration of ethical, social, and economic implications.

## VII. CONCLUSION AND FUTURE WORK

In conclusion, our investigation has unveiled the intricate re- lationships and interconnections between various technolo- gies in the field of artificial intelligence. Through the utiliza- tion of centrality measures, clustering coefficient, and degree of fusion measures, we have gained valuable insights into the fundamental and crucial components that shape the AI land- scape. The results of this study reflect the maturity level of the domain and highlight the key technology areas of fusion in the field. These findings have far-reaching implications for future developments and advancements in AI, providing a clear understanding of the key technology areas that require attention.

Although this research has yielded valuable insights, it is important to acknowledge its limitations. Therefore, future studies should build upon these results by incorporating more dimensions and attributes using the graph bipartite model. This will allow for a more comprehensive understanding of the topic and enable researchers to draw more robust con- clusions. By addressing these gaps in knowledge, future re- search in this field can continue to make important contribu- tions to the advancement of artificial intelligence.

## VIII. ACKNOWLEDGMENT

We acknowledge the support of the AI policy project conducted by CSIR NIScPR, New Delhi. This work was made possible by funding from the National institute of science communication and policy research institute NIScPR. The authors are grateful for their support and contribution to the research.